\documentclass{article}
\usepackage{spconf,amsmath,graphicx}

\usepackage[htt]{hyphenat}

\usepackage{amsmath,amsfonts,bm}

\def\eqref#1{equation~\ref{#1}}

\def\1{\bm{1}}

\def\vv{{\bm{v}}}

\def\mA{{\bm{A}}}
\def\mB{{\bm{B}}}

\def\mS{{\bm{S}}}

\def\mX{{\bm{X}}}
\def\mY{{\bm{Y}}}

\DeclareMathAlphabet{\mathsfit}{\encodingdefault}{\sfdefault}{m}{sl}
\SetMathAlphabet{\mathsfit}{bold}{\encodingdefault}{\sfdefault}{bx}{n}

\newcommand{\R}{\mathbb{R}}

\DeclareMathOperator*{\argmin}{arg\,min}

\usepackage{multicol}
\usepackage{multirow}
\usepackage{makecell}
\usepackage{hyperref}

\usepackage{xcolor}[dvipsnames]
\usepackage[T1]{fontenc}
\usepackage{kpfonts}

\definecolor{joshcolor}{rgb}{0.25, 0.75, 0.55}

\title{On A Guided Nonnegative Matrix Factorization}

\name{Joshua Vendrow$^{\star}$ \qquad Jamie Haddock$^{\star}$ \qquad Elizaveta Rebrova$^{\star}$ \qquad Deanna Needell$^{\star}$\thanks{The authors were partially supported by NSF DMS $\#2011140$ and NSF BIGDATA $\#1740325$.}}
\address{$^{\star}$University of California, Los Angeles\\ Department of Mathematics\\ 520 Portola Plaza, Los Angeles, CA 90095}

\begin{document}
\maketitle
\begin{abstract}
Fully unsupervised topic models have found fantastic success in document clustering and classification.
However, these models often suffer from the tendency to learn less-than-meaningful or even redundant topics when the data is biased towards a set of features.  For this reason, we propose an approach based upon the nonnegative matrix factorization (NMF) model, deemed \textit{Guided NMF}, that incorporates user-designed seed word supervision.  Our experimental results demonstrate the promise of this model and illustrate that it is competitive with other methods of this ilk with only very little supervision information.
\end{abstract}
\begin{keywords}
supervised topic models, supervised nonnegative matrix factorization, seed words
\end{keywords}
\section{Introduction}
\label{sec:intro}
As modern data collection and storage capabilities improve and grow, so do the size and complexity of modern data sets that data practitioners are tasked with turning to actionable knowledge.  For this reason, data scientists are increasingly turning to unsupervised dimensionality-reduction and topic modeling techniques to understand the latent trends within their data.  These approaches have produced fantastic results in document clustering and classification, see e.g.,~\cite{xu2003document,shahnaz2006document}.

However, it has been previously noted that such models can learn topics that are not meaningful or effective in downstream tasks~\cite{chang2009reading}.  In particular, these models can be hindered by data in which certain features are so weighted as to bias the models towards topics with these features and away from more balanced and meaningful topics~\cite{jagarlamudi2012incorporating}.

For this reason, we develop a supervised topic model that incorporates flexible supervision information representing user knowledge of feature importance and associations.  Our approach is based upon the popular nonnegative matrix factorization (NMF)~\cite{lee1999learning} and builds upon its supervised variant, semi-supervised NMF (SSNMF)~\cite{lee2009semi}.  The key difference in our approach, however, is that our goal is to guide the topic outputs, rather than provide labels for classification. The goal is thus to identify topics within the data that are driven by the seeded features, thereby revealing more meaningful topics for the particular application.

\subsection{Nonnegative matrix factorization}
NMF is an approach typically applied in unsupervised tasks such as dimensionality-reduction, latent topic modeling, and clustering.  Given nonnegative data matrix $\mX \in \R_{\ge 0}^{m \times n}$ and a user-defined target dimension $k \in \mathbb{N}$, NMF seeks nonnegative factor matrices $\mA \in \R_{\ge 0}^{m \times k}$, often referred to as the \emph{dictionary} or \emph{topic} matrix, and $\mS \in \R_{\ge 0}^{k \times n}$, often referred to as the \emph{representation} or \emph{coefficient matrix}, such that $\mX \approx \mA \mS$.  There are many formulations of this model (see e.g.,~\cite{cichocki2009nonnegative,lee1999learning,lee2001algorithms}) but the most popular utilizes the Frobenius norm,
\begin{equation}\label{eq:fro_NMF}
\argmin_{\mA\geq 0,\mS\geq 0} \|\mX - \mA \mS\|_F^2.
\end{equation}
Here and throughout, $\mA \ge 0$ denotes the constraint that $\mA$ is entry-wise nonnegative.
The user-defined parameter $k$, which represents the target dimension or the number of believed latent topics, governs the quality of reconstruction of the data; generally $k$ is chosen so that $k < \min\{m,n\}$ to ensure non-triviality of the factorization.  The columns of $\mA$ are often referred to as \emph{topics}; the NMF approximations to the data (columns of $\mX$) are additive nonnegative combinations of these topic vectors.  This property of NMF approximations yields interpretability since the strength of relationship between a given data point (column of $\mX$) and the topics of $\mA$ is clearly visible in the coefficient vector (corresponding column of $\mS$).  For this reason, NMF has found popularity in applications such as document clustering~\cite{xu2003document},
image and audio processing~\cite{guillamet2002non,cichocki2006new},
and financial data mining~\cite{de2008analysis}.

\subsection{Semi-supervised nonnegative matrix factorization}
SSNMF is a modified variant of NMF that jointly factorizes a data matrix $\mX \in \R_{\ge 0}^{m \times n}$ and a supervision information matrix $\mY \in \R_{\ge 0}^{c \times n}$ with the goal of learning a dimensionality-reduction model and a model for a supervised learning task (e.g., classification).  That is, given data matrix $\mX$, supervision matrix $\mY$, and target dimension $k \in \mathbb{N}$, SSNMF seeks the dictionary matrix $\mA \in \R_{\ge 0}^{m \times k}$, representation matrix $\mS \in \R_{\ge 0}^{k \times n}$, and \emph{supervision matrix} $\mB \in \R_{\ge 0}^{c \times k}$ such that $\mX \approx \mA \mS$ and $\mY \approx \mB \mS$.  The most popular SSNMF formulation~\cite{lee2009semi} employs a weighted combination of Frobenius norm terms,
\begin{equation} \label{eq:ssnmf}
   \argmin\limits_{\mA, \mS, \mB\geq 0} \underbrace{\|\mX - \mA \mS\|_F^2}_\text{Reconstruction Error} + \lambda \underbrace{\|\mY - \mB \mS\|_F^2}_\text{Classification Error};
\end{equation}
recently, other formulations have been proposed~\cite{AGHKKLLMMNSW20}.

\subsection{Related work}

Other supervised variants of NMF (besides \eqref{eq:ssnmf}) have been proposed.  The works~\cite{chen2008non, fei2008semi,jia2019semi} propose models that exploit cannot-link or must-link supervision, while~\cite{cho2011nonnegative} introduces a model with information divergence penalties on the reconstruction and on supervision terms that influence the learned factorization to approximately reconstruct coefficients learned before factorization by a support-vector machine (SVM).
Several works~\cite{jia2004fisher,xue2006modified,zafeiriou2006exploiting} propose a supervised NMF model that incorporates Fisher discriminant constraints into NMF for classification.
Joint factorization of two data matrices, like that of SSNMF, is described more generally and denoted Simultaneous NMF in~\cite{cichocki2009nonnegative}.

Previous works incorporating feature-level knowledge into topic modeling have predominantly used as their backbone Latent Dirichlet Allocation (LDA)~\cite{blei2003latent}. The authors of~\cite{andrzejewski2009incorporating} guide topics by incorporating \textit{Must-Links} and \textit{Cannot-Links} that increase or decrease, respectively, the probability of two words appearing in the same topic. The authors of~\cite{andrzejewski2009latent} guide the formation of topics by adding a constraint on the LDA sampling algorithm to force certain words to only appear in specified topics.
Each of the works~\cite{mcauliffe2008supervised,lacoste2009disclda,ramage2009labeled} develop LDA models which incorporate response variables and class labels to improve the learned topic model and its performance on downstream learning tasks.

The work that aligns most closely with our goal is~\cite{jagarlamudi2012incorporating}, which proposes Seeded LDA.  This method accepts sets of seed words and adjusts the LDA model probability distributions to encourage topics to generate words related to those in the seed set. The experimental setup of Seeded LDA is very similar to our own, so within our experiments we provide comparison to Seeded LDA.

Finally, other models utilize other approaches to incorporate feature-level information;
\cite{haghighi2006prototype} utilizes \emph{prototype} supervision information in corpora topic modeling, while that of~\cite{wallach2006topic} utilizes $n$-gram statistics,
while~\cite{griffiths2007topics} is an attempt to extract and utilize \emph{gist} of words in the corpora.
This task is also highly related to that of constrained clustering~\cite{basu2008constrained}.

\subsection{Contribution and Organization}
Our primary contribution is to propose a simple yet worthwhile approach for topic modeling when the user has a priori knowledge about some of the desired topics. For example, we will showcase a setting from political Twitter data where we wish to learn topics related to specific policies and employ seed words to guide the learned topics towards those desired. Without such guidance, the natural topics identified would largely reflect individual political candidates, thereby obscuring the topics of interest and related documents.

With this as our primary objective, in Section \ref{sec:methods} we propose the Guided NMF model and introduce a metric to measure the quality of formed topics. Then, in Section \ref{sec:experiments}, we perform topic modeling experiments on two document analysis data sets and compare our model to Seeded LDA. Finally, in Section \ref{sec:conclusion} we summarize our findings and discuss future work.

\section{METHOD}
\label{sec:methods}

Our proposed method, which we refer to as \emph{Guided NMF}, makes use of seed word (or generally seed-feature) supervision and exploits a model based upon SSNMF.  We evaluate this model on corpora topic modeling and classification tasks.

\subsection{Seed word supervision}
We will refer to a keyword identified in the user-provided supervision as a \textit{seed word}, and we will refer to a (possibly weighted) group of seed words
as a \textit{seed topic}.
We denote a seed topic as a vector $\vv = (v_1, \ldots, v_m)$ (where $m$ denotes the vocabulary size), where
$v_i = 0$ if the $i$th word in the vocabulary is not in the seed topic and some positive weight otherwise.
In general, we expect $\vv$ to be very sparse because the number of important keywords identified for a topic should be far smaller than the total vocabulary of keywords.
 We note that by considering each element of $\vv$ to be a feature rather than a word, we can extend this formulation to any topic modeling task (e.g., in image/video topic modeling tasks). In our experiments, we use $v_i \in \{0, 1\}$ but note that varying weights could improve performance in many applications.

\subsection{Guided NMF} \label{sec:guidednmf}

Let the data matrix $\mX \in \R^{m \times n}$ have examples along the columns and features along the rows and suppose we have seed topics $\vv^{(1)}$, $\vv^{(2)}$, \ldots, $\vv^{(c)} \in R^m$.  Let the \emph{seed matrix} be
\begin{align}
\mY &= [\vv^{(1)}, \vv^{(2)}, \ldots, \vv^{(c)}] \in \R_{\ge 0}^{m \times c}
\end{align}
Guided NMF is formulated as
\begin{equation}
\min \limits_{\mA\geq 0,\mS\geq 0, \mB\geq 0} \|\mX - \mA \mS\|_F^2 + \lambda \|\mY - \mA \mB\|_F.
\end{equation}
 We note that
 this model is symbolically equivalent to standard SSNMF where the data $\mX$ and seed matrix $\mY$ are transposed. Here, the important distinction is the dimension of $\mX$ to which supervision information is provided. This new perspective yields application when there is available information regarding the latent relationship between individual features and topics, rather than individual data points and classes.

Following application of Guided NMF, we can use the topic supervision matrix $\mB$ to identify columns of the dictionary matrix $\mA$ corresponding to the topics that form around our seed words. By examining the corresponding rows of the supervision matrix $\mS$, we can find the documents that Guided NMF attributes to these topics (interpreting the columns of $\mS$ as a score for the relationship of each document to a topic, we can classify documents to the topics based on the magnitude of this score).
To measure accuracy, we use the widely accepted \emph{receiver operating characteristic} (ROC) metric and corresponding \emph{area under the curve} (AUC).
Thus, we use this classification metric as a measure of the quality of topics when they have a one to one correspondence to classes.

\section{EXPERIMENTS}
\label{sec:experiments}
In this section, we present results of applying Guided NMF to 20 Newsgroups and a Twitter political dataset.  We compare with Seeded LDA in the 20 Newsgroups experiments where we have labels. Code for all experiments can be found in \url{https://github.com/jvendrow/GuidedNMF} and uses the multiplicative updates method of~\cite{AGHKKLLMMNSW20}.

\subsection{20 Newsgroups dataset}

The 20 Newsgroups dataset is a collection of approximately 20,000 text documents containing the text of messages from 20 different newsgroups on the distributed discussion system Usenet \cite{KL08}. From this data set, we use a subset of 10 newsgroups with 100 documents each (graphics, hardware, forsale, motorcycles, baseball, medicine, space, guns, mideast, and religion).
In a first example, we consider learning 4 topics but guiding those topics via the
seed words \textit{pitch}, \textit{medical}, and \textit{space} in hopes of capturing the corresponding topics.
In another experiment we use the seed words \textit{motorcycle}, \textit{sale} and \textit{religion} with a similar goal.
We choose rank four so as to capture the topics from the leftover document classes separately, allowing the method to more easily guide the remaining topics as we would hope.
 In Tables \ref{tab:newsgroups_guidednmf_1} and \ref{tab:newsgroups_guidednmf_2}, we display the results of running Guided NMF on the newsgroup dataset with these seed words.
 By including two tables with different seed words, we
 show how the topics vary based on seed information.
 We see that for each seed word, a full topic forms around this word that provides clear and salient keywords corresponding to the information within that class.

\begin{table}[htb]
    \centering
    \caption{Topic keywords learned for a rank 4 Guided NMF on the 20 Newsgroups dataset with the seed words \textit{pitch}, \textit{medical}, and \textit{space}. We see that a clear topic forms from each keyword matching one desired newsgroup class.}
    \begin{tabular}{cccc}
    \Xhline{2\arrayrulewidth}
		\textbf{Topic 1} & \textbf{Topic 2} & \textbf{Topic 3} & \textbf{Topic 4}\\
	\hline
\textit{pitch} & \textit{medical} & \textit{space} & people \\
expected & tests & nasa & know \\
curveball & disease & shuttle & think \\
stiffness & diseases & launch & time \\
loosen & prejudices & sci & use \\
shoulder & services & lunar & new \\
shea & graduates & orbit & see \\
rotation & health & earth & say \\
game & patients & station & us \\
giants & available & mission & god \\
    \Xhline{2\arrayrulewidth}
    \end{tabular}

    \label{tab:newsgroups_guidednmf_1}
\end{table}

\begin{table}[htb]
    \centering
    \caption{Topic keywords learned for a rank 4 Guided NMF on the 20 Newsgroups dataset with the seed words \textit{motorcycle}, \textit{sale}, and \textit{religion}. We see that a clear topic forms from each keyword matching one desired newsgroup class.}
    \begin{tabular}{cccc}
    \Xhline{2\arrayrulewidth}
		\textbf{Topic 1} & \textbf{Topic 2} & \textbf{Topic 3} & \textbf{Topic 4}\\
	\hline
\textit{motorcycle} & \textit{sale} & \textit{religion} & people \\
bike & offer & christian & know \\
dod & condition & judaism & think \\
wheelie & shipping & freedom & time \\
shaft & asking & christians & use \\
bikes & includes & islam & new \\
rider & mb & compulsion & space \\
riding & excellent & avi & see \\
scene & price & life & say \\
ski & best & gunpoint & us \\
    \Xhline{2\arrayrulewidth}
    \end{tabular}

    \label{tab:newsgroups_guidednmf_2}
\end{table}

\subsection{Twitter political dataset}

The Twitter political data set~\cite{DVNPDI7IN_2016} is a data set of tweets sent by political candidates during the 2016 election season. In Table \ref{tab:twitter_nmf}, we display the results of running a regular NMF on the data set. We see that most topics focus on a specific candidate or campaign slogan rather than a political issue.

To uncover ``hidden" topics concerning political issues, we run Guided NMF on this data set with two seed words, \textit{economy} and \textit{obamacare}, two issues discussed during the 2016 election, and display the results in Table \ref{tab:twitter_guidednmf}. We see that a topic forms that around each seed word, and the topic keywords provide additional context for the seeded issue; we see that the main economic concerns are jobs and taxes, and the discussion relating to Obamacare focuses on repeal, for which some Republican candidates advocated.

\begin{table}[htb]
    \centering
    \caption{Topic keywords learned by a rank 8 NMF on the Twitter political dataset. We see that most topics center around one of the political candidates.}
    \begin{tabular}{cccc}
    \Xhline{2\arrayrulewidth}
		\textbf{Topic 1} & \textbf{Topic 2} & \textbf{Topic 3} & \textbf{Topic 4}\\
	\hline
thank & govpencein & gopdebate & tedcruz \\
trump2016 & indiana & imwithhuck & cruz \\
maga$^1$ & indiana\_edc & jeb & cruzcrew \\
great & state & tonight & ted \\
america & jobs & president & choosecruz \\
    \Xhline{2\arrayrulewidth}
    \textbf{Topic 5}& \textbf{Topic 6}& \textbf{Topic 7}& \textbf{Topic 8} \\
    \hline
kasich & hillary & randpaul & fitn \\
john & trump & iowa & new \\
johnkasich & people & iacaucus & hampshire \\
ohio & donald & caucus & johnkasich \\
gov & president & tonight & nh \\
    \Xhline{2\arrayrulewidth}
    \multicolumn{4}{c}{$^1$Here ``maga" abbreviates ``makeamericagreatagain."}
    \end{tabular}

    \label{tab:twitter_nmf}
\end{table}

\begin{table}[htb]
    \centering

    \caption{Topic keywords learned by a rank 8 Guided NMF on the Twitter political dataset with the seed words \textit{economy} and \textit{obamacare}. The first two topics form around these seeds, with meaningful related keywords appearing below them.}

    \begin{tabular}{cccc}
    \Xhline{2\arrayrulewidth}
		\textbf{Topic 1} & \textbf{Topic 2} & \textbf{Topic 3} & \textbf{Topic 4}\\
	\hline
\textit{economy} & \textit{obamacare} & govpencein & gopdebate \\
jobs & fullrepeal & indiana & kasich \\
tax & repeal & indiana\_edc & randpaul \\
plan & replace & state & john \\
create & fight & jobs & tonight \\

    \Xhline{2\arrayrulewidth}
    \textbf{Topic 5}& \textbf{Topic 6}& \textbf{Topic 7}& \textbf{Topic 8} \\
    \hline
tedcruz & hillary & johnkasich & people \\
thank & trump & new & need \\
cruz & donald & fitn & must \\
cruzcrew & clinton & kasich & berniesanders \\
ted & president & hampshire & country \\
    \Xhline{2\arrayrulewidth}
    \end{tabular}

    \label{tab:twitter_guidednmf}
\end{table}

\subsection{Ablation and Comparison}

In all the text-based experiments above, we used only a single seed word per class and achieved salient and interpretable results. Here, we explore the impact of adding additional seed words and also varying the rank of the factorization.
We also provide comparisons to Seeded LDA~\cite{jagarlamudi2012incorporating}. To compute AUC for Seeded LDA, we use the metric described in Section \ref{sec:guidednmf}, but rather than using the $\mS$ matrix as in the case of NMF, we instead use the document-topic distribution variables. For the space topic we use the seed words \textit{space, lunar, nasa, launch, rocket, moon, shuttle} and \textit{orbit} and for the baseball topic we use the seed words \textit{pitch, baseball, team, ball, game, season, base} and \textit{field}.
We choose these seed words from keywords commonly appearing in space or baseball NMF topics. %

In Tables \ref{tab:space_compare} and \ref{tab:baseball_compare}, we display AUC scores for Guided NMF and Seeded LDA over a variety of settings for rank and number of seed words. We see that Guided NMF consistently has an AUC above $0.8$ for all rank and number of seed word choices. In the case of particular interest in our setting, namely when few seed words are supplied and/or only a small number of topics are desired, Guided NMF significantly outperforms Seeded LDA; we note that with a higher rank the desired topics are more likely to form naturally, making the task easier. With many seed words and a high rank, Seeded LDA only slightly outperforms our method. This can likely be attributed to differences between NMF and LDA.
\begin{table}[tb]
    \centering
    \caption{AUC scores for the 20 Newsgroups dataset on documents for the space class.}
    \begin{tabular}{c|c|cccc}
        \Xhline{2\arrayrulewidth}
        \multirow{2}{*}{Rank} & \multirow{2}{*}{Method} &  \multicolumn{4}{c}{\# Seed words} \\
         & & $1$ & $2$ & $4$ & $8$ \\
        \hline

\multirow{2}{*}{4} & Guided NMF & \textbf{0.83} & \textbf{0.88} & \textbf{0.88} & \textbf{0.87} \\
& Seeded LDA & 0.31 & 0.42 & 0.74 & 0.86 \\
\hline
\multirow{2}{*}{6} & Guided NMF & \textbf{0.86} & \textbf{0.87} & 0.88 & 0.87 \\
& Seeded LDA & 0.37 & 0.5 & \textbf{0.91} & \textbf{0.89} \\
\hline
\multirow{2}{*}{10} & Guided NMF & \textbf{0.88} & 0.89 & 0.89 & 0.89 \\
& Seeded LDA & 0.45 & \textbf{0.95} & \textbf{0.95} & \textbf{0.95} \\
\Xhline{2\arrayrulewidth}
    \end{tabular}
    \label{tab:space_compare}
\end{table}

\begin{table}[tb]
    \centering
    \caption{AUC scores for the 20 Newsgroups dataset on documents for the baseball class.}
    \begin{tabular}{c|c|cccc}
        \Xhline{2\arrayrulewidth}
        \multirow{2}{*}{Rank} & \multirow{2}{*}{Method} &  \multicolumn{4}{c}{\# Seed words} \\
         & & $1$ & $2$ & $4$ & $8$ \\
        \hline

\multirow{2}{*}{4} & Guided NMF & \textbf{0.89} & \textbf{0.9} & \textbf{0.9} & \textbf{0.9} \\
& Seeded LDA & 0.31 & 0.42 & 0.74 & 0.86 \\
\hline
\multirow{2}{*}{6} & Guided NMF & \textbf{0.9} & \textbf{0.9} & 0.9 & \textbf{0.9} \\
& Seeded LDA & 0.37 & 0.5 & \textbf{0.91} & 0.89 \\
\hline
\multirow{2}{*}{10} & Guided NMF & \textbf{0.87} & 0.9 & 0.9 & 0.9 \\
& Seeded LDA & 0.45 & \textbf{0.95} & \textbf{0.95} & \textbf{0.95} \\
\Xhline{2\arrayrulewidth}
    \end{tabular}
    \label{tab:baseball_compare}
\end{table}

\section{CONCLUSION}
\label{sec:conclusion}
We propose an NMF-based model, that we call Guided NMF, which incorporates seed topic supervision to guide learned topics towards meaningful and coherent sets of features.  Our initial experiments illustrate the promise of this model in text-based topic modeling applications.  This model could be extended to image/video applications, where the supervision provided encourages object localization and segmentation.

\bibliographystyle{IEEEbib}
\bibliography{bib}

\end{document}